\UseRawInputEncoding
\documentclass[]{interact}
\pdfoutput=1%For ARXIV

\usepackage{epstopdf}% To incorporate .eps illustrations using PDFLaTeX, etc.
\usepackage[caption=false]{subfig}% Support for small, `sub' figures and tables
\usepackage{comment}
\usepackage{makecell}
\usepackage{threeparttable}
\usepackage{booktabs}
\usepackage{multirow}
\usepackage{arydshln}
\usepackage{soul,ulem}
\usepackage[natbibapa,nodoi]{apacite}
\usepackage[dvipsnames]{xcolor}
\theoremstyle{plain}% Theorem-like structures provided by amsthm.sty

\theoremstyle{definition}

\theoremstyle{remark}

\begin{document}

\articletype{ARTICLE TEMPLATE}% Specify the article type or omit as appropriate

\title{FinBERT-MRC: financial named entity recognition using BERT under the machine reading comprehension paradigm}

\author{
\name{Yuzhe Zhang\textsuperscript{a} and Hong Zhang\textsuperscript{a}\thanks{CONTACT Hong Zhang. Email: zhangh@ustc.edu.cn}}
\affil{\textsuperscript{a}School of Management, University of Science and Technology of China, Hefei, China}
}

\maketitle

\begin{abstract}
Recognition of financial entities from literature is a challenging task in the field of financial text information extraction, which aims to integrate a large amount of financial knowledge existing in unstructured texts into structured formats. Implementing financial named entity recognition (FinNER) task under the sequence tagging framework is currently an idiomatic method. However, this method cannot fully take advantage of the semantic information in the texts, which is the bottleneck of such methods. In this work, we formulate FinNER task as a machine reading comprehension (MRC) problem and propose a new model termed FinBERT-MRC. This formulation introduces significant prior information by utilizing well-designed queries, and extracts start index and end index of target entities without decoding modules such as conditional random fields (CRF). We conduct experiments on a publicly available Chinese financial dataset ChFinAnn and a real-word bussiness dataset AdminPunish. Our FinBERT-MRC model achieves average $F_1$ scores of 92.78\% and 96.80\% on the two datasets, respectively, with average $F_1$ gains +3.94\% and +0.89\% over sequence tagging models including BiLSTM-CRF, BERT-Tagger, and BERT-CRF. The source code of FinBERT-MRC is available at \url{https://github.com/zyz0000/FinBERT-MRC}.
\end{abstract}


\section{Introduction}
\label{sec1}
In natural language processing (NLP) field, financial named entity recognition (FinNER) aims to automatically retrieve financial entites (e.g., financial institutions, currencies, and companies) in given texts. It is a prerequisite to precisely and effectively recognize financial entities in order to transform them into knowledge in structured formats. Therefore, the FinNER task has received extensive research attention in recent years. The rule-based symbolic methods has been widely applied in NER tasks for decades \citep{spasic2014text}. About twenty years ago, several approaches based on feature engineering and specific domain knowledge began to appear. Typical representatives of such models used in FinNER task include Hidden Markov Model (HMM) \citep{eddy1996hidden}, Maximum Entropy (ME) \citep{kapur1989maximum}, Support Vector Machine (SVM) \citep{hearst1998support}, Conditional Random Fields (CRF) \citep{lafferty2001conditional}, and Decision Tree (DT) \citep{quinlan1986induction}. Feature engineering not only relies heavily on domain-specific knowledge and handcrafted features, but is also model- and entity-specific. In recent years, the development of neural networks and deep learning has been believed to be a more effective method for NER task \citep{lample2016neural, jagannatha2016structured}. Among them, LSTM-CRF-based methods have gained great popularity on NER task. The vector representations of each word (token) in a sentence are first extracted by LSTM, and are then fed into CRF model for the downstream sequence tagging work. \citet{chiu2016named} used a hybrid network structure by integrating both character-level and word-level features. Their model decoded each tag independently based on a BiLSTM layer followed with a log-softmax layer. \citet{wang2014financial} combined conditional random fields and information entropy to recognize the abbreviation financial named entity candidates. \citet{miwa2016end} proposed a BiLSTM encoder and an incrementally-decoded neural network structure to decode tags jointly. These methods generally encode texts based on recurrent neural network (RNN), but differ in the decoding phase. Very recently, language models (e.g, ELMo \citep{peters-etal-2018-deep}, GPT3 \citep{brown2020language}, and BERT \citep{devlin2018bert}) obtained state-of-the-art (SOTA) performance in many NLP tasks and have gradually become the mainstream in the NLP domain. Compared with the feature engineering methods, deep neural networks are able to automatically extract features and thus can achieve more competitive performance.

The task of NER is commonly formalized as a sequence labeling task: a sequence labeling model \citep{chiu2016named, ma2016end, devlin2018bert} is trained to assign a single tagging class to each unit within a sequence of tokens. However, the models mentioned above, i.e., BiLSTM-CRF and BERT-Tagger, cannot effectively learn the semantic information under the framework of sequence labeling. Inspired by the current trend of formalizing NLP problems into machine reading comprehension (MRC) tasks \citep{levy2017zero, mccann2018natural, li2019entity, li2020unified}, we use FinBERT \citep{yang2020finbert} (namely BERT pretrained on financial corpora) as the backbone in the MRC framework to perform FinNER task, which we believe can enhance the capability of learning the semantic information in financial texts. In the MRC framework, each financial entity type should be encoded by a language query and identified by answering these queries. For example, the task of assigning the \texttt{LOC (LOCATION)} label to “\texttt{So far U.S. soldiers have discovered nearly 600 million dollars hidden around [Baghdad]}” is formalized as answering the query “\texttt{Which location is mentioned in the text?}”. Compared with the sequence labeling framework, the MRC framework has the advantage of introducing prior knowledge, which can contribute to improving model performance.

The remainder of this paper is organized as follows. A brief review of related works is provided in Section \ref{sec2}. Our modeling procedure is described in detail in Section \ref{sec3}. Some quantitative experimental results are presented in Section \ref{sec4}. The paper is concluded with some conclusion remarks in Section \ref{sec5}.

\section{Related Works}
\label{sec2}
\subsection{Named Entity Recognition (NER)}
As one of the basic NLP tasks, NER has been studied extensively ever since it was presented as a sub-task at MUC-6 (Message Understanding Conference) in 1995. In order to complete information extraction from unstructured text,  it is necessary to recognize information units such as Date, Person and Location. Identifying references to these entities in the text is recognized as one of the important sub-tasks of information extraction and is referred to as Named-Entity Recognition and Classification (NERC). The earliest NER tasks were based on lexical features, which are mostly summarized by linguists through study on large corpus and named entity libraries. Rule-based methods apply predefined patterns based on certain features, including keywords, punctuation, and statistical information, to perform string matching. However, these methods are time consuming, expensive, and inflexible. 

To resolve such problems, researchers proposed to treat NER task as a sequence labeling task, i.e., to find the best label sequence for a given input sequence. Typical methods, such as Hidden Markov Models (HMMs), maximum-entropy Markov models (MEMMs), support vector machines (SVMs), and Conditional Random Fields (CRFs), are based on manually-crafted discrete features \citep{yao2009crf, han2011method}. One needs to define handcrafted features that describe the syntactic and semantic essence of the entities to be extracted in the text, before training the model on a sufficient amount of annotated data.

In recent years, neural networks with deep learning have increasingly been used in NER task \citep{lample2016neural}. Neural networks are able to learn features automatically, thereby reducing reliance on handcrafted features compared with other machine learning models. \citet{hammerton2003named} first applied unidirectional LSTMs on NER task. \citet{collobert2011natural} introduced convolutional neural network into NER task and presented the CNN-CRF model. \citet{lample2016neural} combined bidirectional LSTMs with CRFs with features based on character-based word embeddings and unsupervised word embeddings. \citet{ma2016end} and \citet{chiu2016named} used a character-level CNN to extract features from original texts. Thanks to the advent of pre-trained language models such as BERT \citep{devlin2018bert} and ELMo \citep{peters-etal-2018-deep}, the performance of NER is significantly enhanced by the feature extraction capability of pre-trained language models. \citet{devlin2018bert} solved NER task using BERT for token classification model. \citet{gao2021named} proposed a BERT-BiLSTM-CRF model and validated its feature extraction capability on the downstream NER task.

\subsection{Language Models}
The acquisition of word embeddings is a crucial step in NLP tasks. Word embeddings mean learning the latent semantic information of words (tokens) from a large amount of unlabeled data, and map the words (tokens) into dense low-dimensional vectors, which is also called the distributed representation of words \citep{liu2020representation}. In the past decade, several word-embedding techniques have been proposed such as Word2Vec \citep{mikolov2013distributed, 2013Efficient} and GloVe \citep{pennington2014glove}. Word2Vec utilizes the Skip-Gram model to predict the  context words given a target word, or employs the Continuous Bag-Of-Words (CBOW) model to optimize the embeddings so that they can predict a target word given its context words. GloVe uses a weighted least squares model trained on global word-word cooccurence counts. The limitation of Word2Vec and GloVe is also obvious, i.e., word embeddings trained by these methods can only model the context-independent representations due to the existence of slicing windows. 

\begin{figure}[htbp]
\centering 
\includegraphics[width=1\textwidth]{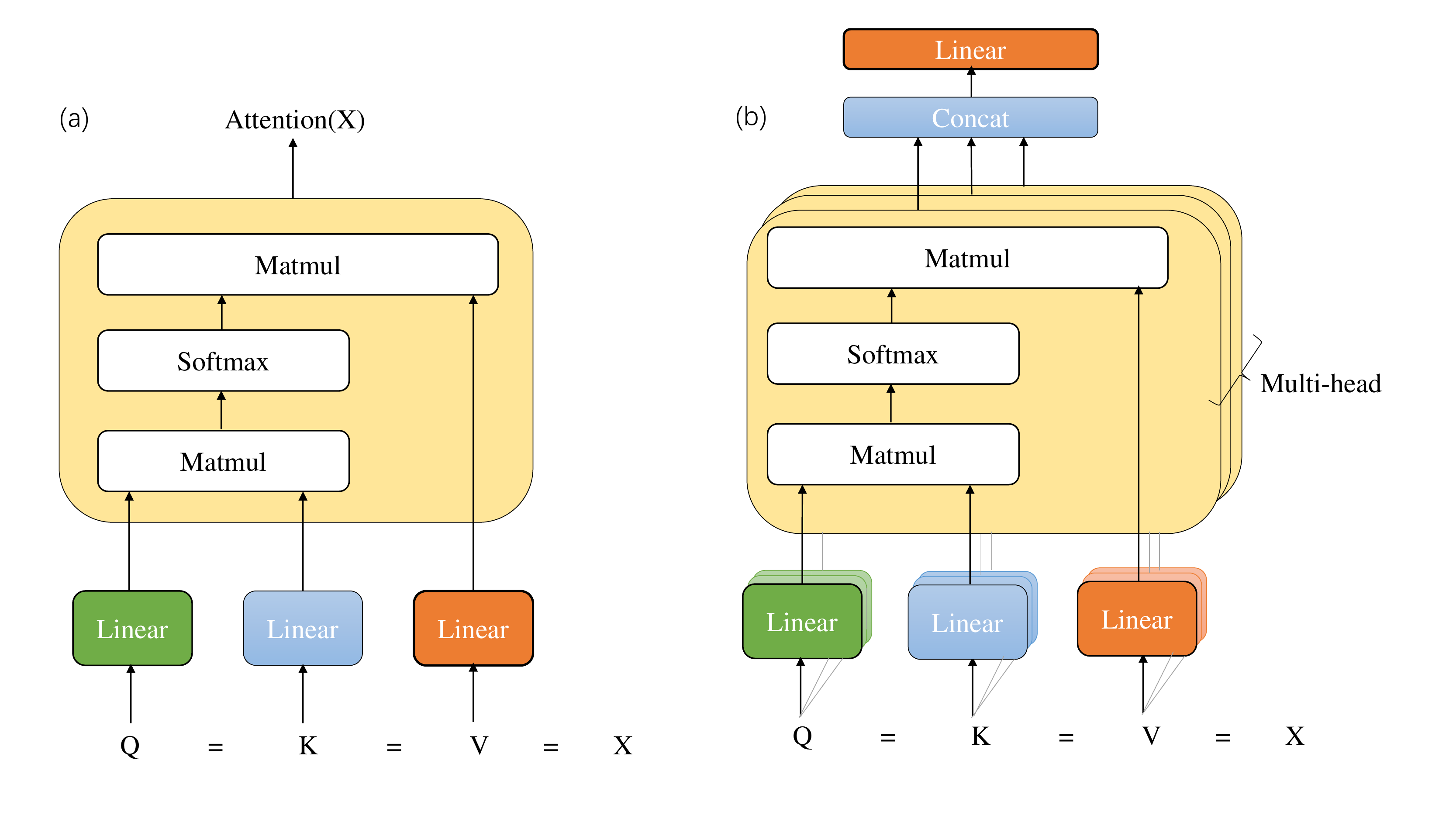}
\caption{(a) Single-head self attention; (b) Multi-head self-attention architecture. For self-attention, the $Q$ (query), $K$ (key), and $V$ (value) should be the same vectors.}
\label{attention}
\end{figure}

The advent of Transformer \citep{vaswani2017attention}, a sequence-to-sequence model based solely on attention mechanisms, brought a significant progress in the field of NLP. The core component of Transformer is the self-attention mechanism, which aims at modeling the strength of relevance between representation pairs, such that a representation can build a direct relation with another representation. Instead of performing single attention function, \citet{vaswani2017attention} proposed multi-head attention to capture different context with multiple individual attention functions. Subfigures (a) and (b) of Figure \ref{attention} show the mechanism of single head attention and multi-head attention. The core operations of single-head attention (denoted as $\text{Attention}$) and multi-head attention (denoted as $\text{MH}$) can be formulated as follows:

$$\text{Attention}(Q, K, V)=\text{softmax}\left(\frac{QK^{\top}}{\sqrt{d_k}}\right)V,$$

$$\text{MH}(Q, K, V)=\text{Concat}(\text{head}_1, \dots, \text{head}_{H})W^{O},$$
where $d_k$ is the dimension of $K$, $\text{head}_{h}=\text{Attention}(QW_{h}^{Q}, KW_{h}^{K}, VW_{h}^{V}), 1 \leq h \leq H$, where $W_{h}^{Q}, W_{h}^{K}, W_{h}^{V}$, and $W^{O}$ are linear transformation matrices with algebraically legal dimensions.

In contemporary natural language processing, pretrained large language models have demonstrated their success in many NLP tasks. Bidirectional Encoder Representations from Transformers (BERT) \citep{devlin2018bert}, which was trained on Wikipedia \footnote{http://wikipedia.org} and BookCorpus \citep{zhu2015aligning}, has become a standard building block for training taks-specific NLP models. It stacks Transformer encoder layers to pretrain representations by jointly conditioning on both left and right context in all layers. Because of the great success of BERT, it has gradually become a mainstream method to utilize pretrained BERT as the feature extractor for input texts and finetune it on the target dataset. 

Pretrained language models using large-scale domain corpora has gradually been studied in recent years, due to the bottleneck that language models pretrained on universal corpora may not suit the domain-specific downstream tasks well. To this end, several domain-specific BERT models are trained and released. BioBERT \citep{lee2020biobert} pretrains a biomedical domain-specific language representation model using large-scale biomedical corpora. SciBERT \citep{beltagy2019scibert} pretrains a scientific domain-specific BERT model using a large-scale scientific publications to improve the performance in downstream scientific NLP tasks. \citet{wang2019smiles} proposed a semi-supervised model named Smiles-BERT to solve different molecular property prediction tasks. \citet{yang2020finbert} are the first to pretrain BERT on financial domain-specific corpora.

\subsection{Machine Reading Comprehension (MRC)}
The aim of MRC models \citep{levy2017zero, mccann2018natural, li2019entity, shen2017reasonet, li2020unified} is to extract answer spans from a context through a given query. The task can be formulated as two classification problems: predicting the start and end positions of the answer spans. Over the past one or two years, there has been a trend of solving NLP tasks using MRC models. \citet{levy2017zero} formulated the task of relation extraction to a question-answering (QA) task: each relation type $R(x, y)$ can be parameterized as a query  $q(x)$ whose answer is $y$. For example, the relation \texttt{ESTABLISHED-IN} can be linked to \texttt{When did x be established?}. Given a query $q(x)$, if a non-null answer $y$ can be returned from a sentence, it means that the relation label for the current sentence is $R(x, y)$. \citet{mccann2018natural} transformed ten different NLP tasks into the QA framework, and all achieved competitive performance. For the NER task, \citet{li2020unified} utilized BERT as the backbone to recognize flat and nested entities from texts under the MRC framework. However, to the best of our knowledge, currently there is no specific research for FinNER under the MRC framework. Our work is inspired by \citet{li2020unified}.

\section{Methodology}
\label{sec3}
\subsection{Task definition}
Given an input sequence $X=\left\{x_1, x_2, \dots, x_n\right\}$, where $n$ is the length of the sequence and $x_i$ is the $i$th word (token) of the sequence. We need to find all entities in $X$, and then assign a label $y \in \mathcal{Y}$ to each entity, where $\mathcal{Y}$ is the predefined set of all possible label types (e.g., $\mathcal{Y} = \left\{\text{LOC, ORG, PER}\right\}$). We formulate the NER task as an MRC task and thus convert the labeling-style NER dataset into a set of (\textit{Question}, \textit{Context}, \textit{Answer}) triplets. Among them, \textit{Context} is the given input sequence $X$, and \textit{Question} is the query sentence designed according to $X$, and \textit{Answer} is the target entity span. For each label type $y \in \mathcal{Y}$, we first construct a question $q_y=\left\{q_1, q_2, \dots, q_m\right\}$ for each sentence, where $m$ equals to the length of the question. An annotated entity $x_{\text{start}, \text{end}}=\left\{x_{\text{start}}, x_{\text{start}+1}, \dots, x_{\text{end}-1}, x_{\text{end}}\right\}$ is a substring of $X$ satisfying $\text{start} \leq \text{end}$. Each entity is associated with a ground-truth label $y \in \mathcal{Y}$. By generating a natural language question $q_y$ based on the label $y$, we can obtain the triplet $(q_y, X, x_{\text{start}, \text{end}})$, which is exactly the triplet (\textit{Question}, \textit{Context}, \textit{Answer}) that we need.

\begin{figure}[htbp]
\centering 
\includegraphics[width=1\textwidth]{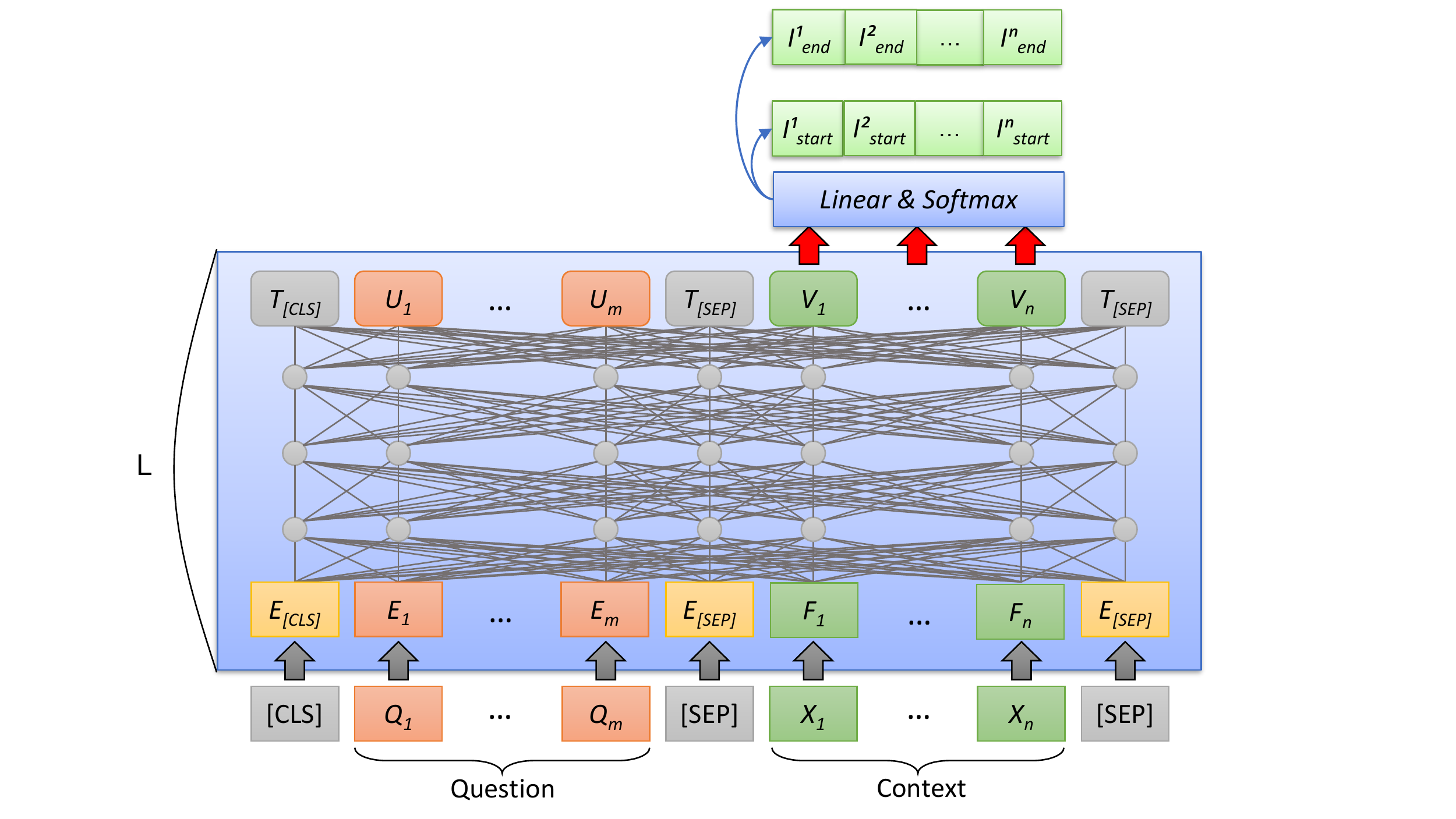}
\caption{The pipeline of FinBERT-MRC consisting of three steps. First, the concatenated question and context is encoded by BERT; Second, the probability of each token to be a start or end index is predicted; Third, the entity spans are selected by applying the nearest matching principle.}
\label{bert}
\end{figure}

\subsection{Query construction}
\begin{figure}[htbp]
	\centering 
	\includegraphics[width=1\textwidth]{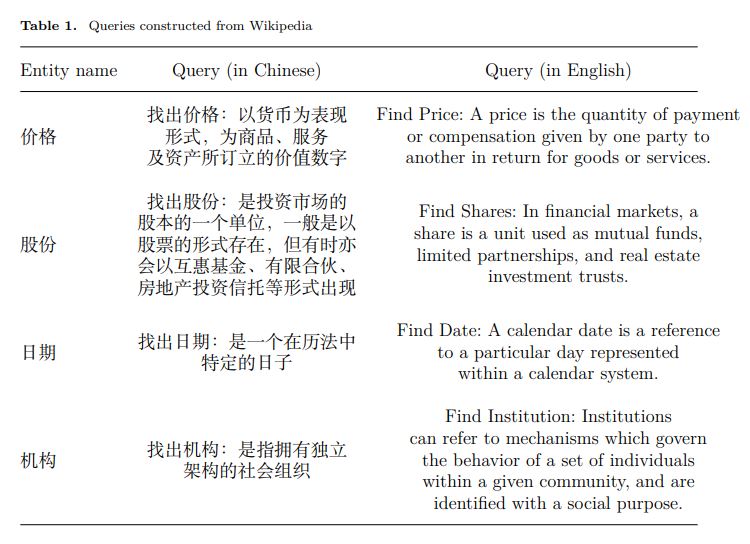}
	\label{queries}
\end{figure}

In the work of \citet{li2020unified}, seven various ways were proposed to construct queries for the entities. In our work, we adopt the scheme which uses the explanation from Wikipedia\footnote{https://zh.wikipedia.org/wiki} to construct queries. Table \ref{queries} lists some examples of the queries we constructed. In addition, we also evaluate the effect of different query construction methods on the model performance in Section \ref{sec4}. 

\begin{comment}
\begin{table}[htbp]
\centering
\caption{Queries constructed from Wikipedia}
\label{queries}
%\resizebox{130mm}{30mm}{
    \begin{tabular*}{\hsize}{@{}@{\extracolsep{\fill}}lcc@{}}
    %\begin{tabular}{ccc}
\toprule
Entity name     & Query (in Chinese)         & Query (in English)    \\
\midrule
\text{价格}       & \makecell[c]{\text{找出价格：以货币为表现} \\ \text{形式，为商品、服务}\\\text{及资产所订立的价值数字}}   & \makecell[c]{Find Price: A price is the quantity of payment \\ or compensation given by one party to    \\ another in return for goods or services.}                                               \\\\
\text{股份}     & \makecell[c]{\text{找出股份：是投资市场的} \\ \text{股本的一个单位，一般是以} \\ \text{股票的形式存在，但有时亦} \\ \text{会以互惠基金、有限合伙、}\\
\text{房地产投资信托等形式出现}} & \makecell[c]{Find Shares: In financial markets, a \\ share is a unit used as mutual funds,  \\limited partnerships, and real estate \\
investment trusts.}                                         \\\\
\text{日期}        & \makecell[c]{\text{找出日期：是一个在历法中} \\ \text{特定的日子}} & \makecell[c]{Find Date: A calendar date is a reference\\ to a particular day represented \\ within a calendar system.}                                                                                                          \\\\
\text{机构}  & \makecell[c]{\text{找出机构：是指拥有独立} \\ \text{架构的社会组织}} & \makecell[c]{Find Institution: Institutions \\ can refer to mechanisms which govern \\ the behavior of a set of individuals \\ within a given community, and are \\identified with a social purpose.} \\
\bottomrule
\end{tabular*}%}
\end{table}
\end{comment}

\subsection{Model formulation}

In this research, we employ FinBERT \citep{yang2020finbert} as our model backbone by considering that it is a domain-specific task. The task diagram is illustrated in Figure \ref{bert}, which consists of three steps.

In the first step, the question $q_y$ and the context $X$ are concatenated, forming the combined input string $$\left\{\text{[CLS]}, q_1, q_2, \dots, q_m, \text{[SEP]}, x_1, x_2, \dots, x_n\right\},$$ where [CLS] and [SEP] are special tokens introduced in the BERT modeling procedure.
Then the combined sequence is fed into FinBERT, which can be depicted by the following formulas:
\begin{align*}
    e_{i}^{0} &= \boldsymbol{W}t_i + b, \\
    e_{i}^{l} &= {Transformer}(e_{i}^{l-1}), \\
    \boldsymbol{E} &= [e_1^{L}, e_{2}^{L}, \dots, e_{n}^{L}]^{\top} \in \mathbb{R}^{n \times d},
\end{align*}
where $t_i$ is the $i$th token embedding, $\boldsymbol{W}$ and $b$ are parameters of linear transformation, $L$ is the number of BERT layers, $l(1 \leq l \leq L)$ is the number of the $l$th BERT layer, and $Transformer$ denotes the Transformer encoder layer, including multi-head attention module, fully connected module, and layer normalization. Finally a context representation matrix $\boldsymbol{E} \in \mathbb{R}^{n \times d}$, where $d$ is the vector dimension of the last Transformer block of BERT and we simply ignore the query representations.

The second and third steps aim to select entity spans. Generally, there are two choices for selecting entity spans in the MRC framework. The first choice is to train two $n$-class classifiers separately to predict the start index and the end index, where $n$ represents the length of the context. Because the Softmax function is calculated on all tokens within the context, this choice has deficiency that each input sentence can only output one span. The second choice is to construct two binary classifiers, one aims to predict whether the token is a start index, and the other serves to predict whether the token is an end index. This strategy allows for outputting multiple start indexes and multiple end indexes for a given context and a specific query, and therefore has the potential to identify all target entities based on $q_y$. In our research, we adopt the second strategy. The detailed description of the second and third steps are as follows:

In the second step, given the representation matrix $\boldsymbol{E}$, the model predicts the probability of each token to be a start index. This procedure is formulated as:
$$\boldsymbol{P}_{\text{start}}=\text{softmax}(\boldsymbol{EW}_{\text{start}}) \in \mathbb{R}^{n \times 2},$$
where $\boldsymbol{W}_{\text{start}}$ is a weight matrix to be learned. Softmax is employed for each row of $\boldsymbol{P}_{\text{start}}$. Similarly, the model predicts the probability of each token to be an end index. The formula is 
$$\boldsymbol{P}_{\text{end}}=\text{softmax}(\boldsymbol{EW}_{\text{end}}) \in \mathbb{R}^{n \times 2},$$
where $\boldsymbol{W}_{\text{end}}$ is a weight matrix to be learned. Softmax is employed for each row of $\boldsymbol{P}_{\text{end}}$.

In the third step, by applying the argmax function to each row of $\boldsymbol{P}_{\text{start}}$ and $\boldsymbol{P}_{\text{end}}$, we predict indexes that might be start or end positions, i.e., 
$$I_{\text{start}}=\left\{i | \text{argmax}(\boldsymbol{P}_{\text{start}}^{i})=1, 1 \leq i \leq n\right\},\ I_{\text{end}}=\left\{j | \text{argmax}(\boldsymbol{P}_{\text{end}}^{j})=1, 1 \leq j \leq n\right\},$$
where the superscripts $i$ and $j$ denote the $i$th and $j$th rows of a matrix, respectively. Because the datasets we conduct experiments on do not contain overlapped entities, we use the nearest matching principle to match the start and end indexes to obtain the predicted entity span. To be specific, if one start (end) index corresponds to multiple end (start) indexes, then only the nearest end (start) index is selected to form the predicted entity span.

\subsection{Training and test}

At the training phase, each context $X$ is paired with two label sequences $Y_{\text{start}},Y_{\text{end}} \in \left\{0, 1\right\}^{n}$, where $Y_{\text{start}}$ is the ground-truth label of each token $x_i$ being the start index of an entity, and $Y_{\text{end}}$ is the ground-truth label of each token $x_i$ being the end index of an entity. We therefore split the loss into two parts:
$$\mathcal{L}_{\text{start}}=\text{CE}(\boldsymbol{P}_{\text{start}}, Y_{\text{start}}),\ \mathcal{L}_{\text{end}}=\text{CE}(\boldsymbol{P}_{\text{end}}, Y_{\text{end}}),$$
where $\text{CE}(\cdot, \cdot)$ denotes the cross-entropy loss function, which can be expressed in the form
$$\text{CE(P, Y)}=-\frac{1}{n} \sum_{i=1}^{n}[y_i \log(p_i) + (1 - y_i) \log(1 - p_i)], $$
where $y_i$ denotes the ground truth label with $y_i=1$ for positive instance and $y_i=0$ for negative instance, $p_i$ is the Softmax probability for the $i$th data point. The training objective function is $$\mathcal{L}_{\text{train}}=\mathcal{L}_{\text{start}} + \mathcal{L}_{\text{end}}.$$

At the test phase, the start and end indexes are first selected based on $I_{\text{start}}$ and $I_{\text{end}}$, respectively. Then, the nearest matching principle is applied to match the start and end indexes, leading to the final extracted answers.

\section{Experiments}
\label{sec4}
\subsection{Experiments on two Chinese FinNER datasets}
\subsubsection{Dataset description}
The first dataset we utilize is a ten-year (2008-2018) ChFinAnn\footnote{Crawling from \url{http://www.cninfo.com.cn/new/index}} documents. This data\sout{ }set is originally aimed at developing algorithms for document-level event extraction (DEE) \citep{zheng2019doc2edag}. This dataset contains 32,040 Chinese financial announcement documents with 24 different types of entities, and it has already been split into training, validation, and test sets. This dataset is available at \url{https://github.com/dolphin-zs/Doc2EDAG/blob/master/Data.zip}. To the best of our knowledge, we are the first to utilize ChFinAnn dataset in the FinNER task research. 

The second dataset is a real bussiness dataset concerning administrative punishment from an App called Enterprise Early Warning developed by Shanghai Financial China Information \& Technology Co.,Ltd, which is referred to as AdminPunish in the remainder of this paper. This dataset collects punishment announcements released by local authorities to punish a citizen, legal representative or organization violating administrative order by reducing rights and interests or increasing obligations. It contains 2,596 annotated sentences with 7 types of entities. The tokens of the sentences are annotated in BIOE format. 

Some summary statistics of the above two datasets are provided in Table \ref{summary}.
\begin{table}[htbp]
\centering
\caption{Summary statistics of two datasets}
\label{summary}
%\begin{tabular}{cccccc}
\begin{tabular*}{\hsize}{@{}@{\extracolsep{\fill}}lccclr@{}}

\toprule
                              & \multicolumn{3}{c}{ChFinAnn}                                               & \multicolumn{2}{c}{AdminPunish} \\
\cmidrule(lr){2-4}  
& Train                      & \multicolumn{1}{c}{Validation}   & \multicolumn{1}{c}{Test}  &                \\
\midrule
No.Docs                              & 25,632                      & \multicolumn{1}{c}{3,204}  & \multicolumn{1}{c}{3,204}  & No.Sents                & 2,596       \\
Avg.Sent/Doc                         & 19.1                      & \multicolumn{1}{c}{22.5} & \multicolumn{1}{c}{21.0} & Avg.SentLength          & 234.9     \\
Avg.Tokens/Doc   & \multicolumn{1}{c}{878.7} & 1077.7                   & 1008.3                   & Avg.EntityNum/Sent      & 6.2        \\
Avg.Entities/Doc  & \multicolumn{1}{c}{10.6}  & 11.3                     & 10.9                     & Avg.EntityLength        & 12.5      \\
Avg.EntityLength & \multicolumn{1}{c}{8.4}   & 8.3                      & 8.4                      & -                       & -          \\
\bottomrule
\end{tabular*}
Abbreviations: No.Docs, number of documents; Avg.Sent/Doc, average number of sentences per document; Avg.Tokens/Doc, average number of tokens per document; Avg.Entities/Doc, average number of entities per document; Avg.EntityLength, average entity length; No.Sents, number of sentences; Avg.SentLength, average sentence length; Avg.EntityNum/Sent, average number of entities per sentence; Avg.EntityLength, average entity length.
\end{table}

\subsubsection{Data preprocessing}
Due to limited computational resources, we cannot use the whole ChFinAnn dataset for model training. Instead, we adopt a resampling scheme to use only a part of the dataset in our experiment. In each sampling replication, we randomly extract 1200, 300, and 300 documents from the original training, validation, and test sets. We delete those sentences that do not contain any entites from the three sets. In order to minimize the variability due to random sampling, we repeat the aforementioned sampling procedure on the ChFinAnn dataset five times. For the AdminPunish dataset, we randomly split it into training, validation, and test sets according to the proportions of 6:2:2. Again, this splitting scheme is repeated five times. 

For ChFinAnn dataset, we properly merge entities according to their actual meanings in order to simplify the problem to some extent. For example, both of the entities ``HighestTradingPrice'' and ``LowestTradingPrice'' represent price, so we integrate them into a common entity ``Price''. This results in 10 types of entities in total for the ChFinAnn dataset. For the AdminPunish dataset, all seven entities remain unchanged. The names of the entities, and their proportions in corresponding datasets are listed in Table \ref{entities}. It can be known that, compared with AdminPunish dataset, entity sparsity exists in ChFinAnn dataset. In ChFinAnn, Institution, Price, and Ratio are the top three least frequent entities, which take very small proportions in the whole dataset.

\begin{table}[htbp]
\centering
\caption{Entity names and the corresponding proportions for two datasets}
\label{entities}
\begin{tabular*}{\hsize}{@{}@{\extracolsep{\fill}}lrlr@{}}
%    \begin{tabular}{cccc}
\toprule
\multicolumn{2}{c}{ChFinAnn} & \multicolumn{2}{c}{AdminPunish} \cr
\cmidrule(lr){1-2} \cmidrule(lr){3-4} 
Entry name & Proportion & Entry name & Proportion\\
\midrule
Price & 3.43\%                  & Client&   27.51\%          \\
Shares &  27.76\%  & FileNumber &  18.33\%  \\
Institution &  0.51\%             & Punishing Authority &  18.65\%          \\
Company &  8.49\%  & Unified Social Credit Codes &  7.82\%     \\
StockAbbr &  8.45\%           & Legal Representative&  11.38\%   \\
StockCode &  9.66\%  &  Id Card Number&   3.41\% \\
Date&   19.77\%  &      Address &  12.90\%     \\
 Ratio &  5.50\% \\
EquityHolder &  10.42\%  \\
Pledgee&   6.00\%       \\
\bottomrule
\end{tabular*}
\end{table}

\subsubsection{Baselines}
We use the following models as baselines:
\begin{itemize}
    \item BiLSTM-CRF from \citet{ma2016end}. BiLSTM-CRF model can be divided into the BiLSTM layer and the CRF layer. The function of the BiLSTM layer is to extract contextual information through input words or word vectors, and determine the probability of any type of label for making prediction. The CRF layer is used to consider the correlation between tags in order to filter the label sequences that are not grammatically reasonable.
    \item BERT-Tagger from \citet{devlin2018bert}. BERT-Tagger treats NER as a tagging task. The model uses BERT as the backbone to extract the representation of input texts, and a followed linear layer transforms the embedding vectors into a lower dimension, the number of tags in the training data. In this way, the NER task is solved in a token classification way.
    \item BERT-CRF from \citet{gao2021named}. BERT-CRF enhances the feature extraction capability using BERT as the feature extractor.
    \item BERT-MRC from \citet{li2020unified}. BERT-MRC unifies the flat and nested named entity recognition under the MRC framework. This formulation naturally tackles flat and nested NER task as extracting entities with different categories requires answering pre-prepared independent questions.
\end{itemize}

\subsubsection{Settings}
For BERT-Tagger, BERT-CRF and BERT-MRC, we utilize the pretrained BERT-base-chinese as the model backbone, and for FinBERT-MRC, we choose FinBERT as the model backbone. Both BERT-base-chinese and FinBERT can be downloaded from  HuggingFace\footnote{https://huggingface.co/} website. 

In order to make fair comparison, we set identical hyperparameters for BERT-MRC and FinBERT-MRC. The hyperparameters used for model training are list in Table \ref{hyperpara}.

The experiments are conducted on a Ubuntu 18.04.5 linux server, with processor Intel(R) Core(TM) i9-7900X CPU @ 3.30GHz, and GPU GeForce RTX 2080 Ti.

\begin{table}[htbp]
\centering
\caption{Hyperparameters for BERT-MRC and FinBERT-MRC training}
\label{hyperpara}
\begin{threeparttable}
\begin{tabular*}{\hsize}{@{}@{\extracolsep{\fill}}lccccc@{}}
%\begin{tabular}{lccccc}
\toprule
Model             & seqlen & lr   & bs & epochs & loss     \\
\midrule
BiLSTM-CRF        & -      & 1e-3 & 64 & 50     & NLL \\
BERT-Tagger       & 512    & 5e-5 & 4  & 10     & CE       \\
BERT-CRF          & 512    & 5e-5, 5e-3 & 32 & 10     & NLL \\
BERT(FinBERT)-MRC & 512    & 5e-5 & 8  & 10     & CE      \\
\bottomrule
\end{tabular*}
%\begin{tablenotes}
\footnotesize
Abbreviations: seqlen, the maximum sequence length; bs, the batch size; lr, the learning rate; NLL, the negative log likelihood loss; CE, the cross-entropy loss; epochs, the minimum number of training epochs when both BERT-MRC and FinBERT-MRC reaches convergence.

In the BERT-CRF model, the learning rate of CRF layer is set to be $5 \times 10^{-3}$ and those in the other part of the model are set to be $5 \times 10^{-5}$. This trick aims to accelerate the convergence of CRF layer.
%\end{tablenotes}
\end{threeparttable}
\end{table}

\subsubsection{Results}
The performance of each model is evaluated with the $F_1$-score ($F_1$) defined as $F_1=2PR / (P + R)$, where $P$ and $R$ are precision and recall, respectively. We report the mean value of each measure ($P$, $R$, $F_1$) as well as the standard deviation on the test sets in mean$\pm$std format. The test results on two datasets are listed in Table \ref{ChFinAnn} and Table \ref{CxkChar}. Detailed experimental results can be found in Figure \ref{ChFinAnn_res} and Figure \ref{CxkChar_res} in Appendices.

Evidently, FinBERT-MRC achieves the best $F_1$ scores on both ChFinAnn and AdminPunish datasets, with average $F_1$ scores of 92.78\% and 96.80\%, respectively. For these two datasets, FinBERT-MRC has $F_1$ score gains of 0.88\% and 0.23\%, respectively, over BERT-MRC. This demonstrates that using language models pretrained on domain-specific corpora effectively enhances the performance on the downstream tasks. Besides, FinBERT-MRC has $F_1$ score gains of +2.75\% and +0.27\% over BERT-CRF on the ChFinAnn and AdminPunish datasets, respectively. This greater performance boost could be due to the fact that some entities of ChFinAnn are sparse since MRC formulation is more immune to the tag sparsity issue.

\begin{table}[htbp]
\centering 
\caption{Prediction accuracy measures (mean$\pm$std) on test sets of the ChFinAnn dataset}
\label{ChFinAnn}
%\begin{tabular}{cccc}
\begin{tabular*}{\hsize}{@{}@{\extracolsep{\fill}}lccc@{}}
\toprule
    Model           & Precision & Recall & $F_1$ score \\
    \midrule
    BiLSTM-CRF      & $87.89 \pm 1.80$ &$90.19 \pm 0.97$ & $89.00 \pm 0.73$ \\
    BERT-Tagger     & $85.54 \pm 0.61$  & $90.23 \pm 1.09$  & $87.75 \pm 0.39$   \\
    BERT-CRF        & $90.80 \pm 1.50$  & $89.35 \pm 2.68$  & $90.03 \pm 0.89$  \\
    BERT-MRC        & $91.78 \pm 0.76$  & $92.15 \pm 0.60$  & $91.90 \pm 0.42$    \\
    FinBERT-MRC     & $\bold{92.31 \pm 0.46}$  & $\bold{93.33 \pm 0.79}$ &  $\bold{92.78 \pm 0.56}$\\
\bottomrule
\end{tabular*}
\end{table}

\begin{table}[htbp]
\centering
\caption{Prediction accuracy measures (mean$\pm$std) on test sets of the AdminPunish dataset}
\label{CxkChar}
\begin{tabular*}{\hsize}{@{}@{\extracolsep{\fill}}lccc@{}}
\toprule
    Model           & Precision & Recall & $F_1$ score \\
    \midrule
    BiLSTM-CRF      & $96.36 \pm 0.39$ & $96.62 \pm 0.57$ & $96.49 \pm 0.43$ \\
    BERT-Tagger     & $93.19 \pm 1.33$  & $96.34 \pm 1.09$  & $94.71 \pm 0.88$   \\
    BERT-CRF        & $\bold{96.58 \pm 0.37}$  & $96.48 \pm 0.77$  & $96.53 \pm 0.46$  \\
    BERT-MRC        & $96.21 \pm 0.25$  & $96.95 \pm 0.41$  & $96.57 \pm 0.23$    \\
    FinBERT-MRC     & $96.45\pm 0.53$  & $\bold{97.16 \pm 0.29}$ &  $\bold{96.80 \pm 0.38}$\\
\hline
\end{tabular*}
\end{table}

\subsection{Ablation Studies}
\subsubsection{The effect of query design strategy}
In the machine reading comprehension field, the construction of query is known to have some impact on the model performance. Intuitively, the more information a query contains, the better performance a model can achieve. In this subsection, we explore the effect of the following three different query construction methods:
\begin{description}
\item[Keyword] A query is the keyword describing the tag. 
\item[Rule-based query template] This generates queries using templates. 
\item[Wikipedia] A query is constructed using its Wikipedia definition. 
\end{description}
It is worth mentioning that there are other query construction methods, such as synonyms, keyword+synonyms, and so on. Nevertheless, the reason why they are not studied in our experiment is that they might not be appropriate for the ChFinAnn and AdminPunish datasets. For example, it is difficult to find synonyms for financial entity "legal representative". Synonyms may be more suitable for universal entities such as organization and location. We hold the view that it accomodates to the FinNER datasets in our experiment by constructing query using keyword, rule-based query template, and Wikipedia.  

\begin{figure}[htbp]
  \centering
    \includegraphics[width=15cm, height=5cm]{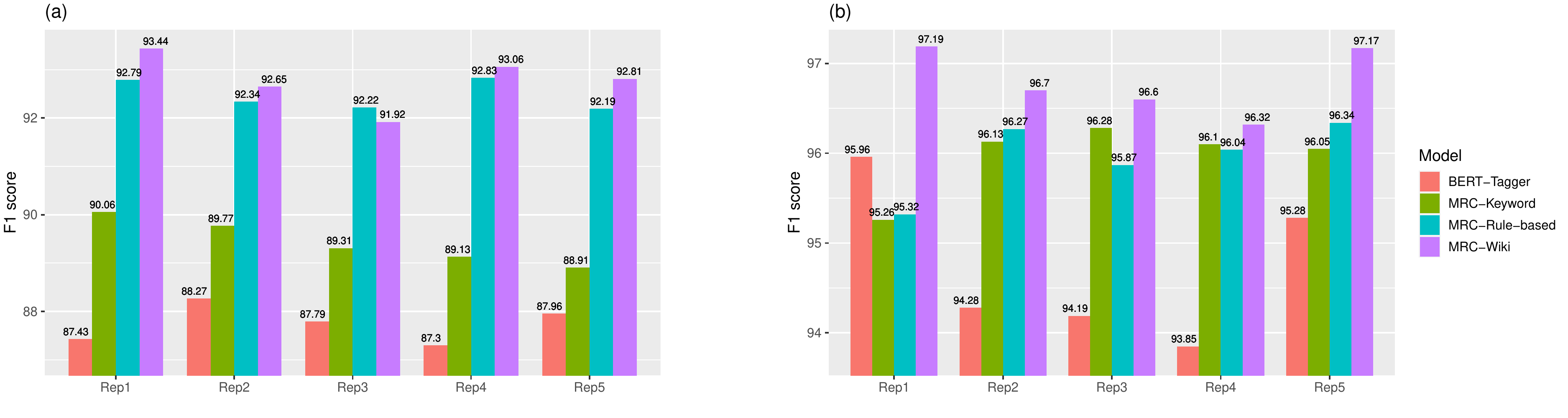}    
    \caption{$F_1$ scores of four query construction methods in five replications of experiments. (a) The ChFinAnn dataset; (b) The AdminPunish dataset.}
     \label{query_res}
\end{figure}

In this way, we can comprehensively analyze the effect of different query construction methods on FinBERT-MRC. We follow the same hyperparameter setting in Table \ref{hyperpara} and change the queries to train FinBERT-MRC on the ChFinAnn and AdminPunish datasets. Table \ref{question} summaries the $F_1$ scores of three query construction methods: Keyword (MRC-Keyword), Rule-based query template (MRC-rule-based) and Wikipedia (MRC-Wiki), as well as the baseline method BERT-Tagger. Evidently, All query construction methods outperform the baseline method BERT-Tagger on the two datasets. This conforms with the intuition that incorporates prior knowledge can boost prediction performance. Figure \ref{query_res} showcases the $F_1$ scores of various models. MRC-Wiki is shown to performs the best in all replications, while MRC-Keyword and MRC-Rule-based outperform BERT-Tagger expcet in one replication for the AdminPunish dataset. 

\begin{table}[htbp]
\centering
\caption{$F_1$ scores with various types of queries}
\label{question}
\begin{tabular*}{\hsize}{@{}@{\extracolsep{\fill}}lccc@{}}
\toprule\multicolumn{2}{c}{ChFinAnn} & \multicolumn{2}{c}{AdminPunish} \cr
\cmidrule(lr){1-2} \cmidrule(lr){3-4} 
Model (query) &$F_1$ &Model &$F_1$ \cr
\midrule
BERT-Tagger &  $87.75 \pm 0.39$                 & BERT-Tagger &  $94.71 \pm 0.88$          \\
MRC-Keyword &   $89.44 \pm 0.47$          & MRC-Keyword &  $95.96 \pm 0.40$          \\
MRC-Rule-based &  $92.47 \pm 0.31$          & MRC-Rule-based & $95.97 \pm 0.41$\\
MRC-Wiki &  $\bold{92.78 \pm 0.56}$                    & MRC-Wiki        & $\bold{96.80 \pm 0.38}$  \\
\hline
\end{tabular*}
\end{table}

%\begin{figure}[htbp]
%  \centering
%  \begin{minipage}[t]{.4\linewidth}
%    \includegraphics[width=5cm, height=5cm]{pics/ChiFinAnn_F_query.pdf}
%  \end{minipage}
%  \begin{minipage}[t]{.4\linewidth}
%    \includegraphics[width=7cm, height=5cm]{pics/CxkChar_F_query.pdf}
%  \end{minipage}
%    \caption{The $F_1$ score using different query construction method. (a) results on ChFinAnn dataset; (b) results on AdminPunish dataset.}
%     \label{query_res}
%\end{figure}

\subsubsection{The effect of training sample size}

\begin{figure}[htbp]
  \centering
    \includegraphics[width=12cm, height=5cm]{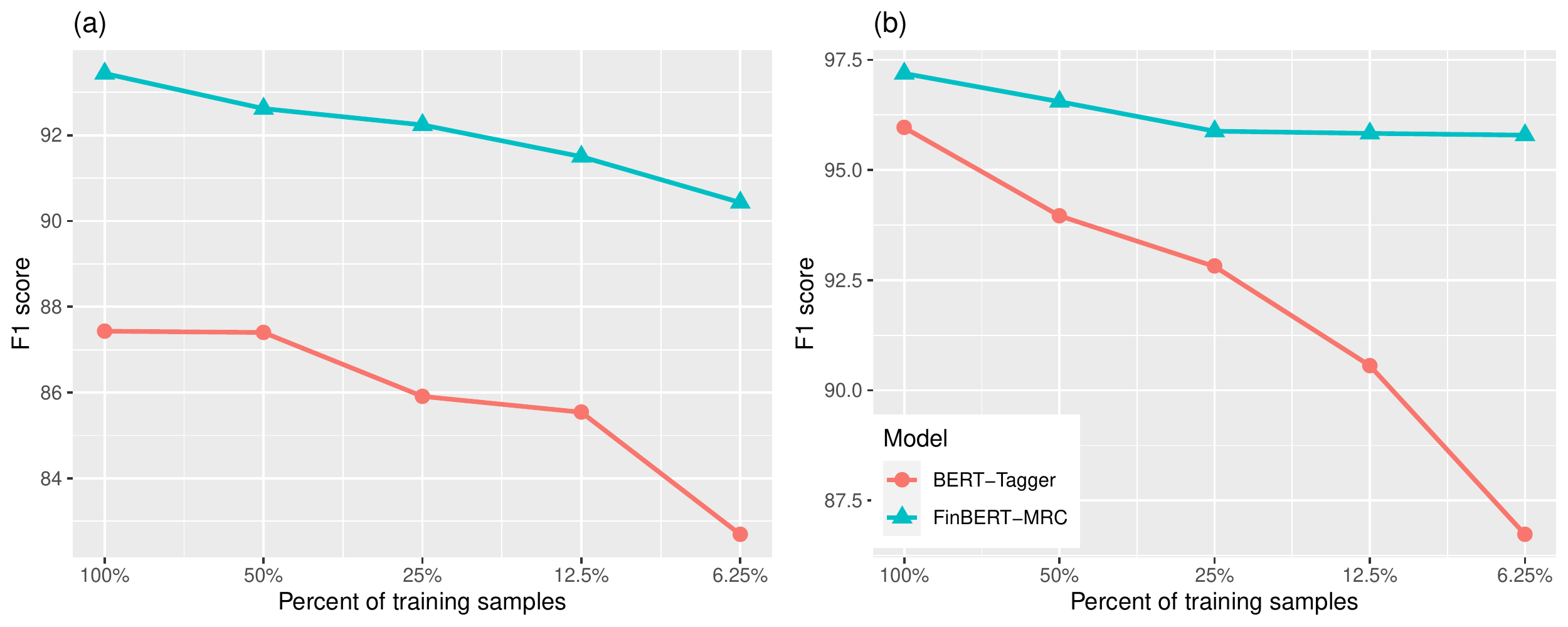}
    \caption{The effect of training sample size on $F_1$ score. (a) The ChFinAnn dataset; (b) The AdminPunish dataset.}
     \label{sample_size}
\end{figure}

We also the impact of the size of training samples on the performance of FinBERT-MRC. %Since the queries related to entities encodes prior knowledge, we expect that FinBERT-MRC can still work well with less training data. 
Figure \ref{sample_size} shows $F_1$ scores on test samples with various sizes of training samples. For both BERT-Tagger and FinBERT-MRC, the $F_1$ score is decreasing with the training sample size, but FinBERT-MRC appears to be more robust, further demonstrating the superiority of utilizing the entity encode prior knowledge for FinBERT-MRC. 
%\begin{figure}[htbp]
%  \centering
%  \begin{minipage}[t]{.45\linewidth}
%    \includegraphics[width=6cm, height=5cm]{ACM/pics/size_ChiFinAnn.pdf}
%  \end{minipage}
%  \begin{minipage}[t]{.4\linewidth}
%    \includegraphics[width=6cm, height=5cm]{ACM/pics/size_CxkChar.pdf}
%  \end{minipage}
%    \caption{Effect of varying percentage of training samples on (a) ChFinAnn dataset, (b) AdminPunish dataset. FinBERT-MRC can achieve the same $F_1$-score comparing to BERT-Tagger with fewer training samples.}
%     \label{sample_size}
%\end{figure}

\subsection{Several Representative Examples}

\begin{figure}[htbp]
	\centering
	\includegraphics[width=15cm, height=8cm]{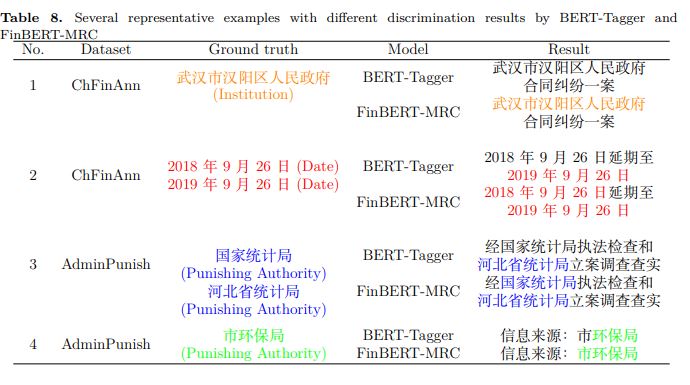}    
	\label{case}
\end{figure}

In Table \ref{case}, we present several representative examples with different discrimination results between BERT-Tagger and FinBERT-MRC. In the first example, FinBERT-MRC correctly identifies the entity (an institution) while BERT-Tagger misses it. This example shows that FinBERT-MRC has advantage over BERT-Tagger in terms of learning syntactic information under the context of financial text field. In the second and third examples, BERT-Tagger does not recognize all entities of the same type in one sentence, while FinBERT-MRC can help alleviate this problem by sufficiently learning the whole sentence. In the fourth example, BERT-Tagger only identifies a fragment of the true entity, while FinBERT-MRC correctly obtains the complete entity. This phenomenon reflects that FinBERT-MRC can better distinguish the boundary of entities. 

The above four examples veries the effectiveness of FinBERT-MRC in the syntactic learning, showing that using domain-specific pretrained language model can achieve a better generalization capability. Through this case study, we can infer that FinBERT-MRC outperforms BERT-Tagger because of its superiority in syntactic and semantic learning. To be specific, FinBERT-MRC can correct some false negative instances in the presence of multiple entities in one sentence, and can alleviate the boundary recognition problem to some extent.

\section{Discussion and conclusion}
\label{sec5}
In this work, we use BERT to perform FinNER task under the MRC framework. Compared with the conventional methods which solve NER task under the sequence labeling framework, using FinBERT in the MRC framework can enhance the ability for identifying target entities of the model. We think that the improvement is due to two aspects. First, the application of pretrained language model trained on domain-specific corpora learns better semantic representation for input texts, which is more beneficial for domain-specific downstream tasks. Second, the appropriately designed queries encode important prior knowledge about the entity to extract and the prior knowledge is learned jointly with the original texts. This mechanism instructs the model to find the entity span accurately. The proposed model demonstrates its effectiveness in the real-word bussiness dataset. It is of great interest to explore the effectiveness of low resource, few-shot NER task under the MRC framework, which deserves further investigation.

\section*{Acknowledgement}

We thank Shanghai Financial China Information \& Technology Co.,Ltd for providing the AdminPunish dataset for this research.

%\section*{Disclosure statement}

%An unnumbered section, e.g.\ \verb"\section*{Disclosure statement}", may be used to declare any potential conflict of interest and included \emph{in the non-anonymous version} before any Notes or References, after any Acknowledgements and before any Funding information.

\section*{Funding}

The work of YZ and HZ is partially supported by the National Natural Science Foundation of China (No. 72091212) and the Anhui Center for Applied Mathematics.

%\section*{Notes on contributor(s)}

%An unnumbered section, e.g.\ \verb"\section*{Notes on contributors}", may be included \emph{in the non-anonymous version} if required. A photograph may be added if requested.

%\section*{Nomenclature/Notation}

%An unnumbered section, e.g.\ \verb"\section*{Nomenclature}" (or \verb"\section*{Notation}"), may be included if required, before any Notes or References.

%\section*{Notes}

%An unnumbered `Notes' section may be included before the References (if using the \verb"endnotes" package, use the command \verb"\theendnotes" where the notes are to appear, instead of creating a \verb"\section*").

%\bibliographystyle{apacite}
%\bibliography{sample}

\newpage
\section*{Appendices}

\subsection*{Detailed experiment results on the ChFinAnn and AdminPunish datasets}
%\begin{figure}[htbp]
%  \centering
%  \begin{minipage}[t]{.3\linewidth}
%    \includegraphics[width=4.3cm, height=5cm]{pics/ChiFinAnn_P.pdf}
%  \end{minipage}
%  \begin{minipage}[t]{.3\linewidth}
%    \includegraphics[width=4.3cm, height=5cm]{pics/ChiFinAnn_R.pdf}
%  \end{minipage}
%  \begin{minipage}[t]{.3\linewidth}
%    \includegraphics[width=6cm, height=5cm]{pics/ChiFinAnn_F.pdf}
%  \end{minipage}
%    \caption{The performance comparison of five models on ChFinAnn dataset: (a) precision; (b) recall; (c) $F_1$ score.}
%     \label{ChFinAnn_res}
%\end{figure}

\begin{figure}[htbp]
  \centering
    \includegraphics[width=15cm, height=5cm]{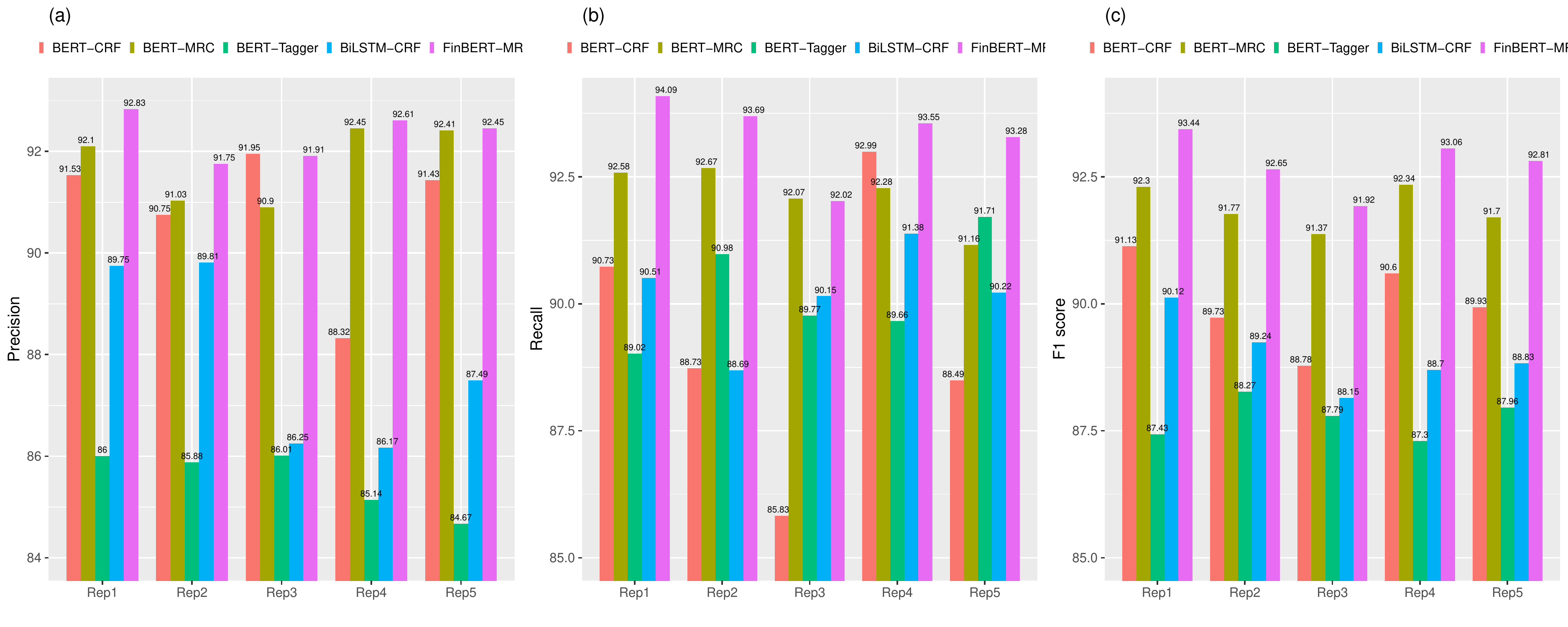}    
    \caption{Prediction accuracy measures of five models on the ChFinAnn dataset based on five replications of experiments. (a) Precision; (b) Recall; (c) $F_1$ score.}
     \label{ChFinAnn_res}
\end{figure}

%\begin{figure}[htbp]
%  \centering
%  \begin{minipage}[t]{.3\linewidth}
%    \includegraphics[width=4.3cm, height=5cm]{pics/CxkChar_P.pdf}
%  \end{minipage}
%  \begin{minipage}[t]{.3\linewidth}
%    \includegraphics[width=4.3cm, height=5cm]{pics/CxkChar_R.pdf}
%  \end{minipage}
%  \begin{minipage}[t]{.3\linewidth}
%    \includegraphics[width=6cm, height=5cm]{pics/CxkChar_F.pdf}
%  \end{minipage}
%    \caption{The performance comparison of five models on AdminPunish dataset: (a) precision; (b) recall; (c) $F_1$ score.}
%     \label{CxkChar_res}
%\end{figure}

\begin{figure}[htbp]
  \centering
    \includegraphics[width=15cm, height=5cm]{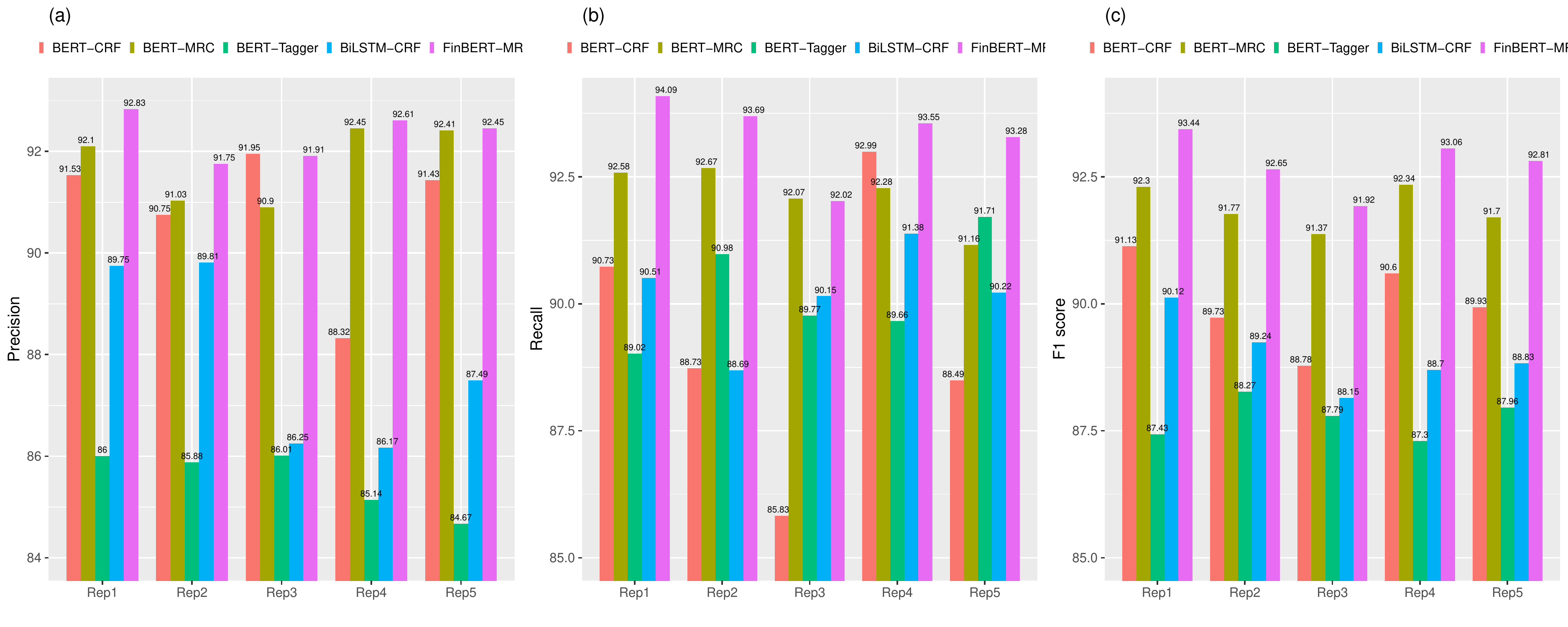}    
    \caption{Prediction accuracy measures of five models on the AdminPunish dataset based on five replications of experiments. (a) Precision; (b) Recall; (c) $F_1$ score.}
     \label{CxkChar_res}
\end{figure}

\end{document}